%% file: main.tex
\documentclass[sigconf,screen]{acmart}

\usepackage[utf8]{inputenc}
\usepackage[T1]{fontenc}
\usepackage{microtype}
\usepackage[ruled,vlined]{algorithm2e}
\usepackage{pbox}

\usepackage{amsmath,xcolor,pifont}
\usepackage{multirow}
\usepackage{multicol}
\usepackage{balance}

\usepackage{amssymb}% http://ctan.org/pkg/amssymb
\usepackage{url}

\newcommand{\cmark}{\textcolor{blue}{\ding{51}}}%
\newcommand{\xmark}{\textcolor{red}{\ding{55}}}%

\newtheorem{definition}{Definition}

\newcommand{\sys}{ModelDiff\xspace} % name of the system
\newcommand{\bench}{ModelReuse\xspace} % name of the benchmark
\newcommand{\benchN}{114\xspace} % number of models in the benchmark
\newcommand{\benchNP}{144\xspace} % number of models in the benchmark
\newcommand{\benchNPdirect}{84\xspace} % number of direct-reuse models in the benchmark
\newcommand{\benchNPcombine}{60\xspace} % number of combined-reuse models in the benchmark
\newcommand{\correctness}{91.7\%\xspace} % correctness achieved

\newcommand{\etal}{\textit{et al}.~}
\newcommand{\ie}{\textit{i}.\textit{e}.~}
\newcommand{\eg}{\textit{e}.\textit{g}.~}

\DeclareMathOperator*{\argmax}{arg\,max}

\DeclareMathOperator*{\mean}{\textit{mean}}
\DeclareMathOperator{\diversity}{\textit{diversity}}
\DeclareMathOperator{\divergence}{\textit{divergence}}
\DeclareMathOperator{\similarity}{\textit{sim}}
\DeclareMathOperator{\dist}{\textit{dist}}
\DeclareMathOperator{\XP}{\textit{XP}}
\DeclareMathOperator{\DDV}{\textit{DDV}}

\setcopyright{acmcopyright}
\acmPrice{15.00}
\acmDOI{10.1145/3460319.3464816}
\acmYear{2021}
\copyrightyear{2021}
\acmSubmissionID{issta21main-p203-p}
\acmISBN{978-1-4503-8459-9/21/07}
\acmConference[ISSTA '21]{Proceedings of the 30th ACM SIGSOFT International Symposium on Software Testing and Analysis}{July 11--17, 2021}{Virtual, Denmark}
\acmBooktitle{Proceedings of the 30th ACM SIGSOFT International Symposium on Software Testing and Analysis (ISSTA '21), July 11--17, 2021, Virtual, Denmark}

\begin{document}
\title{\sys: Testing-Based DNN Similarity Comparison for Model Reuse Detection} % TODO: replace with your title

\author{Yuanchun Li}
% \authornotemark[2]
% \email{trovato@corporation.com}
% \orcid{1234-5678-9012}
% \author{G.K.M. Tobin}
% \authornotemark[1]
\affiliation{%
  \institution{Microsoft Research}
  \streetaddress{5 Danling Street}
  \city{Beijing}
  \country{China}
}
\email{Yuanchun.Li@microsoft.com}

\author{Ziqi Zhang}
% \authornote{Work done while Yunxin Liu was working at Microsoft and Ziqi Zhang was an intern at Microsoft. Ziqi Zhang contributed equally as a co-primary author. $\dagger$Yuanchun Li is the corresponding author.}
\affiliation{%
  \institution{Peking University}
  \streetaddress{5 Yiheyuan Road}
  \city{Beijing}
  \country{China}}
\email{ziqi_zhang@pku.edu.cn}

\author{Bingyan Liu}
\affiliation{%
  \institution{Peking University}
  \streetaddress{5 Yiheyuan Road}
  \city{Beijing}
  \country{China}}
\email{lby_cs@pku.edu.cn}

\author{Ziyue Yang}
\affiliation{%
  \institution{Microsoft Research}
  \streetaddress{5 Danling Street}
  \city{Beijing}
  \country{China}
}
\email{Ziyue.Yang@microsoft.com}

\author{Yunxin Liu}
% \authornotemark[1]
\affiliation{%
  \institution{Institute for AI Industry Research (AIR), Tsinghua University}
%   \streetaddress{30 Shuangqing Rd}
  \city{Beijing}
%   \state{Beijing}
  \country{China}
}
\email{liuyunxin@air.tsinghua.edu.cn}

%%
%% By default, the full list of authors will be used in the page
%% headers. Often, this list is too long, and will overlap
%% other information printed in the page headers. This command allows
%% the author to define a more concise list
%% of authors' names for this purpose.
% \renewcommand{\shortauthors}{Li et al.}

\begin{abstract}
The knowledge of a deep learning model may be transferred to a student model, leading to intellectual property infringement or vulnerability propagation. Detecting such knowledge reuse is nontrivial because the suspect models may not be white-box accessible and/or may serve different tasks.
In this paper, we propose \sys, a testing-based approach to deep learning model similarity comparison. Instead of directly comparing the weights, activations, or outputs of two models, we compare their behavioral patterns on the same set of test inputs. Specifically, the behavioral pattern of a model is represented as a decision distance vector (DDV), in which each element is the distance between the model's reactions to a pair of inputs. The knowledge similarity between two models is measured with the cosine similarity between their DDVs.
To evaluate \sys, we created a benchmark that contains \benchNP pairs of models that cover most popular model reuse methods, including transfer learning, model compression, and model stealing. Our method achieved \correctness correctness on the benchmark, which demonstrates the effectiveness of using \sys for model reuse detection. A study on mobile deep learning apps has shown the feasibility of \sys on real-world models.
% The knowledge of a deep learning model may be transferred to a student model, leading to intellectual property infringement or vulnerability propagation. Detecting such knowledge reuse is nontrivial because the suspect models may not be white-box accessible and/or may serve different tasks.
% In this paper, we propose ModelDiff, a testing-based approach to deep learning model similarity comparison. Instead of directly comparing the weights, activations, or outputs of two models, we compare their behavioral patterns on the same set of test inputs. Specifically, the behavioral pattern of a model is represented as a decision distance vector (DDV), in which each element is the distance between the model's reactions to a pair of inputs. The knowledge similarity between two models is measured with the cosine similarity between their DDVs. To evaluate ModelDiff, we created a benchmark that contains 144 pairs of models that cover most popular model reuse methods, including transfer learning, model compression, and model stealing. Our method achieved 91.7% correctness on the benchmark, which demonstrates the effectiveness of using ModelDiff for model reuse detection. A study on mobile deep learning apps has shown the feasibility of ModelDiff on real-world models.
\end{abstract}

% % TODO: replace this section with code generated by the tool at https://dl.acm.org/ccs.cfm
% \begin{CCSXML}
% <ccs2012>
% <concept>
% <concept_id>10002978.10003029.10011703</concept_id>
% <concept_desc>Security and privacy~Usability in security and privacy</concept_desc>
% <concept_significance>500</concept_significance>
% </concept>
% </ccs2012>
% \end{CCSXML}



% \ccsdesc{Security and privacy~Use https://dl.acm.org/ccs.cfm to generate actual concepts section for your paper}
% % -- end of section to replace with generated code

% \keywords{Adversarial attack; membership inference; trojan attack; fairness; transfer learning; model compression} % TODO: replace with your keywords


\maketitle
% \begin{figure}
% \centering
% \includegraphics[width=\linewidth]{figure/xxx.pdf}
% \caption{xxx.}
% \label{figure:xxx}
% \end{figure}

% \begin{table}
% \begin{tabular}{cc}
% \toprule
% Table & template \\
% \midrule
% \bottomrule
% \end{tabular}
% \caption{xxx.}
% \label{table:xxx}
% \end{table}

\input{sec_introduction}

\input{sec_background}

\input{sec_approach}

\input{sec_evaluation}
\input{sec_related_work}

\input{sec_limitations}

\input{sec_conclusion}

% \section*{Acknowledgments}
% The USENIX latex style is old and very tired, which is why
% there's no \textbackslash{}acks command for you to use when
% acknowledging. Sorry.

% \section*{Availability}

% USENIX program committees give extra points to submissions that are
% backed by artifacts that are publicly available. If you made your code
% or data available, it's worth mentioning this fact in a dedicated
% section.
% \nocite{*}

\bibliographystyle{ACM-Reference-Format}
\balance
\bibliography{reference}

\end{document}

%% file: sec_introduction.tex
\section{Introduction}

Deep learning models (\ie deep neural networks, or DNNs for short) are increasingly deployed into various applications for a wide range of tasks. Due to the difficulty of building accurate and efficient models from scratch, various model reuse techniques have been proposed to help developers build models based on existing models.
The knowledge of an existing model can be transferred to new models that are tailored for different application scenarios and/or resource constraints. For example, transfer learning \cite{pan2009survey} can be used to adapt the existing models trained for one task to solve other similar tasks. Model compression techniques \cite{han2015deep} can convert a large model to a smaller one to deploy in resource-constrained environments while reserving reasonable accuracy.
Due to the great convenience and remarkable performance, these techniques are increasingly used by deep learning developers today.

% Model reuse is problematic
However, the ability of knowledge transfer also leads to concerns about intellectual property (IP) and vulnerability propagation.
First, a deep learning model is usually an important property for a company given the difficulty of training it \cite{coleman2017dawnbench}. Reusing a model without authorization or license compliance would violate the IP right. 
% The recently-found model extraction attacks \cite{tramer2016stealing,orekondy2019knockoff} make the problem even more severe since attackers can steal a black-box model by continuously querying it.
Second, some pretrained models may have security defects (such as adversarial vulnerability \cite{szegedy2013intriguing}, backdoors \cite{trojannn,li2021deeppayload}, etc.), and the models based on them may inherit the defects \cite{davchev2019empirical_adv_transfer,yao2019latent}.
Similar problems exist in traditional programs where the code may be plagiarized or reused, and software similarity analysis \cite{rattan2013software_clone_review,ragkhitwetsagul2018comparison,yao2018redundancy} is one of the most popular techniques to address such problems.

% Challenges of model similarity
Analyzing the similarity between deep learning models involves three key challenges.
First, the models under comparison, especially the suspect models built upon pretrained models are usually not white-box accessible, since many of them are deployed on a server and provided to customers through inference APIs.
Second, even if the models are available for structure or weight comparison, the structural similarity does not necessarily mean knowledge similarity:
% unlike code reuse that can be detected based on the similarity between program structures (control-flow graphs, data-flow graphs, abstract syntax trees, etc.), 
Two unrelated DNNs may have identical or similar structures, since they may use the same public state-of-the-art model architecture (\eg ResNet, MobileNet, etc.).
Meanwhile, two closely related DNNs may have significantly different structures and weights, for example when one is generated from another through knowledge distillation.
Third, the models that contain common knowledge may appear quite different since they may use the knowledge for different tasks (\eg an object detection model built upon an image classification model through transfer learning).

\begin{figure}
    \centering
    \includegraphics[width=8.5cm]{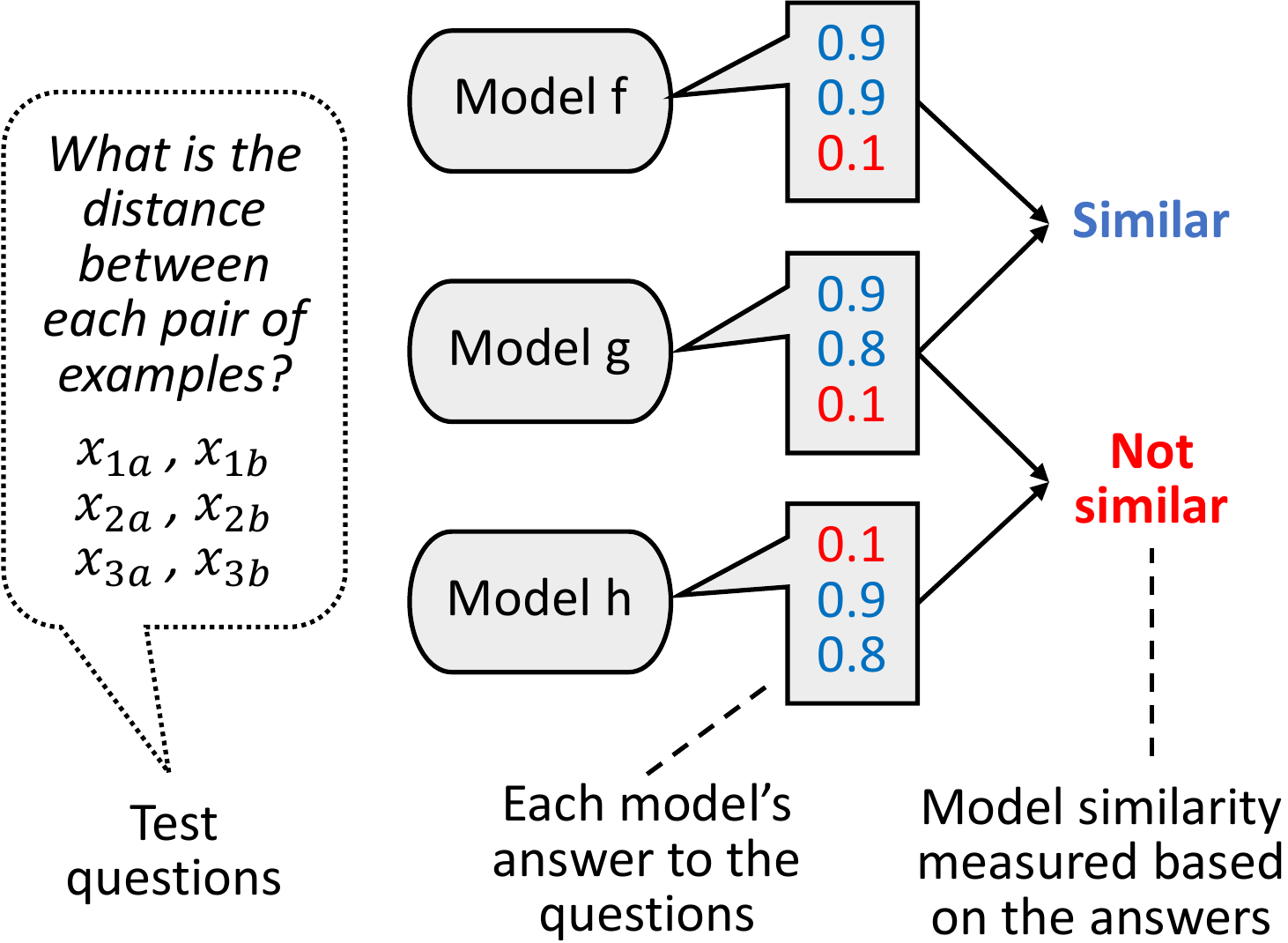}
    \caption{An illustration of the idea of \sys to measure knowledge similarity between DNN models.}
    \label{fig:illustration}
\end{figure}

% Our solution
In this paper, we propose \sys, a testing-based approach to DNN model similarity comparison. Instead of directly comparing the graph structures and weights that may be unavailable or incomparable, we compare the decision patterns of the models based on how they respond to the same set of test inputs.
Measuring knowledge similarity from the testing perspective directly solves the first two challenges stated above since it only requires black-box access to the suspect models and no comparison of the internal structures is needed.
However, due to the third challenge (the models may belong to different tasks), the test outputs of different models are not directly comparable.
Thus we introduce a new data structure, named decision distance vector (DDV), to represent the decision logic of a model on the test inputs. Each value in a DDV is the distance between the outputs of the model produced by two inputs.
The insight behind DDVs is that two models would group the test inputs with a similar pattern if their decision boundaries are similar.
Since the size of a DDV is only related to the number of inputs used to test the model, the DDVs generated with the same set of samples are comparable across different models.
As a result, the knowledge similarity between DNNs can be measured based on the distance between their DDVs.
An illustration of the idea is shown in Figure~\ref{fig:illustration}.

To detect model reuse, a key problem of \sys is how to generate the test inputs that can represent the unique decision pattern of a model that is exclusively shared by the models built upon it.
Using normal samples as test inputs is ineffective because the normal inputs are usually processed by the common-sense knowledge that is shared across unrelated models.
For example, feeding a normal cat image to different image classifiers would lead to similar outputs, although the classifiers might be trained with completely different datasets and algorithms.

Inspired by prior work on adversarial attack transferability \cite{demontis2019why_transfer,huang2019intermediate_attack,davchev2019empirical_adv_transfer}, we use both normal and adversarial inputs to construct the test inputs. The insight is that the normal inputs and adversarial inputs are processed by the normal and imperfect knowledge of a model respectively, while the decision boundary of the model can be characterized by the combination of normal and imperfect knowledge.
Specifically, given two models under comparison, one of them is selected as the target model and another is the suspect model. The test inputs in \sys are generated based on a set of normal samples (named seed inputs) that lie in the input distribution of the target model.
We find an adversarial input for each seed input by maximizing the divergence between the model's predictions on adversarial and normal inputs and the diversity of the model predictions produced by the test inputs.
Each adversarial input and the corresponding normal input are paired to compute DDVs that depict the decision boundary precisely and completely.
Such DDVs can capture the similarity between teacher and student model because the decision boundaries are transferred during model reuse.

To evaluate our approach, we created a benchmark named \bench. \bench contains \benchN models generated from large pretrained models using various model reuse techniques. Based on these models, we obtained \benchNP pairs of models that have reused knowledge, including \benchNPdirect direct reuses (one is generated from another using a single model reuse method) and \benchNPcombine combined reuses (one is generated from another using a combination of transfer learning and model compression).
We evaluated \sys by examining whether it can be applied to detect these reuses (feasibility) and whether it can correctly compute higher similarity scores for the model pairs with reused knowledge (correctness).

\sys could support meaningful comparison for all model pairs in \bench benchmark and achieved an overall correctness of \correctness, which outperformed both the white-box and black-box baseline methods that we created based on weight, feature map, and fingerprint comparison.
% In most cases, there is clear gap between the similarity scores of reused model pairs and irrelevant model pairs, meaning that a threshold can be found to detect model reuse.
% Meanwhile, \sys is able to compare 100\% of the model pairs, including the model pairs with different tasks or different structures.

To better understand the knowledge similarity measured with \sys, we further analyzed the relation between the similarity score and the model accuracy.
% and vulnerability transferability.
The result shows that the similarity score computed by \sys is in general proportional to model accuracy, \ie a higher similarity between two models typically means more useful knowledge of one model is utilized by another, leading to higher test accuracy.
% while more vulnerabilities of one model are transferred to another as well.

Finally, to examine whether \sys can be used to measure model similarity in the wild, we collected 35 TFLite models from 20,000 real-world Android apps in Google Play and compared them with a popular pretrained model using \sys. Our method was able to handle these real-world black-box models, and the knowledge similarities measured for these models were consistent with our manual inspection based on the model file names.

This paper makes the following key contributions:

\begin{enumerate}
    \item To the best of our knowledge, this is the first work that systematically discusses the problem of DNN model reuse detection, where the student model and teacher model may be heterogeneous, black-box, and serving different tasks.
    \item We introduce a benchmark named \bench for model reuse detection, which contains \benchNP models generated with popular model reuse techniques with varying configurations.
    \item We propose \sys, a testing-based method for model similarity comparison. Our method achieved \correctness correctness on \bench benchmark. Both the benchmark and the tool will be released to the community.
\end{enumerate}

%% file: sec_background.tex
\section{Background: DNN Model Reuse}

Building an efficient and accurate DNN model from scratch is a data-intensive and time-consuming task, thus it is common for developers to build DNNs based on existing pretrained DNNs.
This section will introduce transfer learning and model compression, two widely-used techniques for adapting a DNN model to different tasks and different resource constraints, and model stealing, a malicious way to transfer the knowledge of a model.
We call the models being reused as \emph{teacher} models and the models that inherent knowledge from teacher models as \emph{student} models.

\subsection{Transfer Learning}

Transfer learning aims to transfer the knowledge of a pretrained teacher model to a student model used for a different but related problem. For example, an image classifier that predicts the type of animals in the input images can transfer knowledge to a more specific classifier that predicts the breeds of cats, or to an object detector that predicts the location of each animal in the image.
The reason why transfer learning is feasible is that DNNs trained for similar tasks usually share a common feature extraction process. For example, in computer vision, DNNs usually try to detect edges in the earlier layers, shapes in the middle layer, and some task-specific features in the later layers, thus the early and middle layers can be shared for different tasks.
Transfer learning was systematically summarized by Pan~\etal~\cite{pan2009survey}. 
% Later researchers empirically dig into the transferability of DNN features~\cite{azizpour2015factors,yosinski2014transferable}.
% Azizpour~\etal investigated several critical factors that affect input representation transferability~\cite{azizpour2015factors}. Yosinski~\etal systematically qualify the generality and specificity of each layer~\cite{yosinski2014transferable}.
Today, transfer learning is widely used in computer vision and natural language processing tasks today thanks to the rapid advance of pretrained models in these areas.
% Many machine-learning-as-a-service (MLaaS) platforms such as Amazon ML Services, Microsoft Azure ML, and Google Cloud AI also offer transfer learning services to facilitate developers in building models from pretrained models. 

The most straightforward method to implement transfer learning is fine-tuning. To fine-tune a model, developers first replace the last layer of the teacher model with a customized layer whose output shape is tailored for the developers' task. Then the last few layers in the new model are retrained with the (mostly small-scale) training data in the application scenario. The weights in other layers are fixed or slightly adjusted during retraining so that most knowledge in the teacher model is preserved. Fine-tuning is also the recommended way to implement transfer learning in the tutorials of most deep learning frameworks.

% Thus, in transfer learning, we only need to retrain the latter layers, which greatly reduces the requirements of training data and computing power.

\subsection{Model Compression}

Model compression is used to shrink a DNN model so that it can be deployed to devices with limited storage memory and/or computation ability, such as smartphones, smart cameras, and vehicular systems.
The main techniques to implement model compression include model quantization, pruning, and knowledge distillation.

\textbf{Weight Quantization} compresses model size and speeds up inference by quantizing model weights to low-bit value \cite{han2015deep}.
A common practice is to cut model weights from 32-bit floating-point values to 8-bit integer values. Specifically, the floating-point weight on one layer is scaled and shifted to an integer range and the decimals are clipped. During inference, the weight is recovered by the scale factor and shift factor and participates in the computation. 
% During quantization, the high-precision values are truncated as they have less influence on the output value compared to low-precision majorities. 
% Han~\etal first introduced the idea of quantization and demonstrate the feasibility on the ImageNet dataset. It was later transferred to small devices such as FPGA~\cite{gudovskiy2017shiftcnn} and integrated with other techniques to achieve a better performance~\cite{zhou2017incremental}. 
% Besides traditional usage, quantization was also integrated with adversarial defense to improve network advantage~\cite{lin2019defensive, rakin2018defend}. Our work differs with them in the broad types of defect that is not limited to adversarial examples. And we focus on the defect heritage phenomenon instead of harden the quantization process.

\textbf{Model Pruning} shrinks the model by slimming less-important parts. There are two major pruning methods, including weight pruning and channel pruning. Weight pruning~\cite{han2015learning} means to cut weight connections by setting the weights to zero, which can lead to higher computation speed with sparse matrix-based acceleration.
Channel pruning~\cite{li2016pruning} refers to cut less-important output channels of convolution layers to reduce the number of weights.
Today, the typical pipeline (also recommended in the tutorials of popular deep learning frameworks) of model pruning involves three steps, including training, weight pruning, and fine-tuning.

% The idea of weight pruning was first proposed by Han~\etal to ease the intensive use of computation and storage in the deep learning~\cite{han2015learning}. They introduced a training-pruning-fine-tuning pipeline that is widely used in today's implementation. Li~\etal measure the channel importance by L1-norm of the convolution weight and prune less important filters~\cite{li2016pruning}. The importance of pruned model weights is recently studied by Liu~\etal and Frankle~\etal at the same time~\cite{liu2018rethinking, frankle2018lottery}, but they arrive at different conclusions. In this paper, we consult to the traditional method and use the pruned model weight to initialize fine-tuning stage. 

\textbf{Knowledge Distillation} \cite{hinton2015distilling} transfers model knowledge by using the intermediate features and outputs of the teacher model to train the student model. Ideally, the student model can achieve comparable performance with the teacher model but with a much smaller size and faster speed. Unlike transfer learning which transfers knowledge from one task to another, knowledge distillation requires the teacher and student to have the same label space (\ie the same task). Compared with other model compression methods, knowledge distillation has the flexibility to customize the student model architecture.

% Hinton~\etal first proposed to distill the knowledge via prediction probability from an ensemble of models to a single model. Later, researchers concentrate on the transmission media that is used to transfer more knowledge: Romero~\etal use ~\cite{romero2014fitnets} inter-media features and Yin~\etal utilized flow between layers~\cite{yim2017gift}. 
% Despite improving student model accuracy, KD can also defend against adversarial examples~\cite{papernot2016distillation}.

\subsection{Model Stealing}

An adversary can also steal the knowledge with only black-box access to the teacher model \cite{tramer2016stealing,orekondy2019knockoff}. For example, most machine-learning-as-a-service (MLaaS) platforms provide prediction APIs instead of the whole models. An attacker can obtain training data by continuously querying the prediction APIs, then use the training data to train a new model. This method is similar to knowledge distillation, while less effective in transferring knowledge since the intermediate features are unused.
% Later work attempts to improve stealing efficiency by introducing better input selection strategies to query the prediction API \cite{}.

\section{Motivation and Goal}
\label{sec:motivation_and_goal}

We are motivated by two issues related to model reuse, intellectual property infringement and vulnerability propagation. 

\textbf{Intellectual Property Infringement.} An accurate and efficient DNN model is an important property of a company since it involves much intellectual effort and computing resources. Training AlphaGo Zero from scratch costs around 35 million dollars in computing power \cite{dan:how_much}, and a recent NLP model is estimated to cost about 4.6 million dollars to achieve the best accuracy \cite{li:openai:review}. The total cost of inventing, building, and testing the models would be much higher. Unauthorized reuse (\eg using models protected by non-commercial licenses for commercial purposes) or theft of such models would be a severe violation of the IP rights \cite{guo2018watermarking,cao2019ipguard}.
% Although most public models today are 

\textbf{Vulnerability Propagation.} DNN models are found to be vulnerable against various types of attacks, such as adversarial attacks \cite{szegedy2013intriguing,goodfellow2014explaining_adversarial} that can generate inputs that can lead to prediction errors and backdoor attacks \cite{gu2017badnets,trojannn,li2021deeppayload} that can control the output of a model by injecting specific hidden logic into it. Recent studies have found that these vulnerabilities are transferable \cite{davchev2019empirical_adv_transfer}, \ie if the teacher model has vulnerabilities that are known to attackers, the student model may inherit the defective logic that can easily be exploited by the attackers. The transferability can be further improved by tailoring more advanced attacks \cite{wang2018great_training_vulnerability,yao2019latent,huang2019intermediate_attack,dong2019evading,rezaei2019target_agnostic_attack}. Once a pretrained model is found vulnerable or malicious, it is important to find and notify the apps based on the problematic model.

The two issues are widely-discussed and well-understood in traditional software. For example, it is well-known that reusing third-party software and open-source libraries may be subject to code reuse licenses, and reusing buggy or vulnerable libraries may lead to severe security incidents.
% understanding the license of open-source software is necessary before any commercial use, otherwise the developers may be charged for license violations.
% There are also many serious incidents caused by 
% , including the recently-reported HeartBleed bug in OpenSSL library. 
Code reuse is an important topic in software engineering research, and code similarity analysis \cite{rattan2013software_clone_review,ragkhitwetsagul2018comparison} is one of the most widely-used techniques to deal with such problems.

% However, how to analyze DNN model similarity has rarely been discussed. 
As DNN models are rapidly gaining popularity and increasingly used as core components in many software applications, we anticipate that the IP and vulnerability propagation issues of DNNs may also become non-negligible in the future. This motivates us to investigate the problem of \emph{DNN knowledge similarity comparison}:

\begin{definition}
\textbf{(DNN knowledge similarity comparison)}
Given two DNN models, the goal of DNN knowledge similarity comparison to compute a similarity score estimating how likely one model is built upon another using model reuse techniques such as transfer learning, model compression, etc.
\end{definition}

We assume that models under comparison have the same input shape and similar input statistical distribution (\eg both models accept RGB images as inputs), which is true for most model reuse methods today. In fact, two models would unlikely to have similar knowledge if they deal with different types of inputs.

We identify three related challenges in DNN similarity analysis.

\begin{enumerate}
    \item \textbf{Black-box Access.} Unlike the program code that can be decompiled from the applications for comparison, DNNs, especially the suspect student DNNs are usually hosted on a server or compressed into an irreversible format for better accuracy and efficiency. Thus utilizing the internal structures or intermediate representations for comparison is sometimes infeasible.
    \item \textbf{Model Heterogeneity.} Even if the models are white-box available, a student model built with the teacher model may prune parameters or completely change the structure during transfer learning or compression. Meanwhile, the similarity between model structures does not mean knowledge similarity, \eg some state-of-the-art model architectures are open-sourced and used by different developers for diverse tasks.
    \item \textbf{Task Difference.} Comparing models based on their black-box inference APIs is also not straight-forward as the models may serve for different tasks. For example, a student model may reuse the knowledge of an animal classifier to classify medical images via transfer learning. Thus directly comparing the model outputs may not be feasible.
    % \item \textbf{Uninterpretable logic.} Model interpretability has always been a major limiting factor of DNNs. The decision of a DNN is made with millions of numerical operations that look meaningless to humans. The inability to understand the decision logic poses a unique challenge when comparing the knowledge between models - even experienced experts cannot identify knowledge similarity by inspecting the decision process.
    % \item \textbf{Large model size.} Even though some model reuse methods would not significantly change the parameters, detecting such similarity is still hard due to the large model size. Directly comparing the weight vectors is infeasible if the student model developers simply reorder the DNN nodes. Instead, identifying same (or close) weights between two DNNs is a graph isomorphism problem, which is time-consuming to solve given the fact that state-of-the-art DNN models typically have millions even billions of weights.
\end{enumerate}

These challenges make it impossible to adopt most traditional code similarity analysis techniques that are based on graph (control-flow graph, data-flow graph, abstract syntax tree, etc.) comparison for model similarity analysis.

%% file: sec_approach.tex
\section{Our Approach: \sys}

We propose a testing-based method named \sys for DNN knowledge similarity comparison. The key idea is to interpret the knowledge of a DNN with its reaction to a set of test inputs that can be represented as a decision distance vector (DDV). The similarity between models can be measured by comparing their DDVs computed for the same set of inputs.
Instead of the model structures, weights, and outputs that are difficult or unavailable to compare, the DDVs of different models are uniform and easy to obtain, which addresses the challenges mentioned in Section~\ref{sec:motivation_and_goal}.
The symbols that will be commonly used in this paper are shown in Table~\ref{table:symbols}.

\begin{table}
\centering
\caption{Definition of symbols commonly used in this paper.}
\begin{tabular}{cc}
\toprule
Symbol & Meaning \\
\midrule
% $\mathcal{M}=(\mathcal{N}, \mathcal{S})$ & \tabincell{c}{Model $\mathcal{M}$ with neuron set $\mathcal{N}$ and synapse set $\mathcal{S}$} \\
$f$, $g$, $h$ & DNN models under comparison \\ \hline
$f \sim g$ & One of $f$ and $g$ is reused from another \\ \hline
$X, x_i$ & Input set $X$ with $i$-th element $x_i$ \\ \hline
$\XP, x_i, x_i'$ & A list of input pairs $\XP$ with $i$-th pair $x_i, x_i'$ \\ \hline
$f(x_i)$ & Output of model $f$ produced by input $x_i$ \\ \hline
% $f^{act}(x_i)$ & Intermediate neuron values of $f$ activated by $x_i$ \\ \hline
$\similarity(f, g)$ & Knowledge similarity between model $f$ and $g$ \\ \hline
$\dist(a, b)$ & Distance between vector $a$ and $b$ \\ \hline
% $\diversity_f(X)$ & Diversity of outputs of $f$ produced by inputs in $X$ \\ \hline
$\DDV_f$ & Decision distance vector of model $f$ \\
\bottomrule
\end{tabular}
\label{table:symbols}
\end{table}

\subsection{Approach Overview}

\begin{figure*}
    \centering
    \includegraphics[width=14cm]{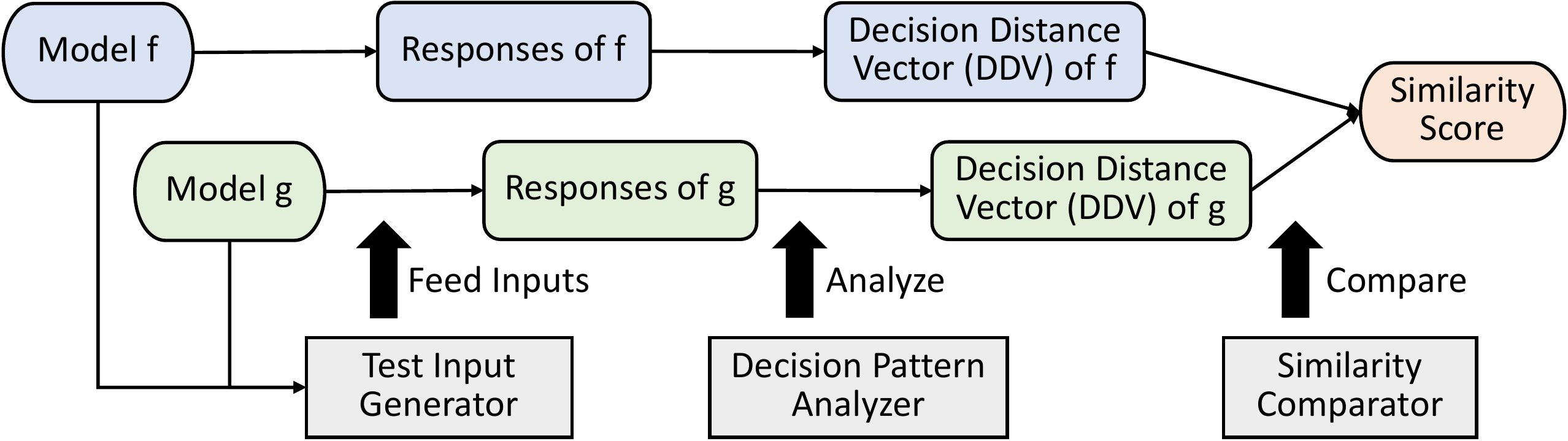}
    \caption{The pipeline of \sys to measure knowledge similarity between two models.}
    \label{fig:pipeline}
\end{figure*}

The pipeline of knowledge similarity analysis in \sys is shown in Figure~\ref{fig:pipeline}.
The main components of \sys include a test input generator, a decision pattern analyzer, and a vector similarity comparator.

Given two DNN models $f$ and $g$, the test input generator first generates several input examples $X$ that can trigger diverse reactions in the models, then the input examples are grouped into pairs $\XP$ fed into the $f$ and $g$ one by one. For each input example $x \in X$, we record the responses of both models to the input as $f(x)$ and $g(x)$. The overall decision logic of a model is represented as a decision distance vector ($\DDV_f$ and $\DDV_g$ for $f$ and $g$ respectively).
Finally, the similarity between the two models is measured by computing the distance between their $\DDV$s.
The following subsections will explain the key components in detail.

\subsection{Test Input Generation}
\label{sec:approach:test_input_generation}

The goal of test input generation is to create an input dataset that can capture the decision logic shared by DNN models that contain reused knowledge.
% The reactions should cover as many behavioral patterns of the model as possible to generate thorough results.

Given two models $f$ and $g$ under comparison, we first select one model (say $f$) as the target model, and another model ($g$) is the suspect model. In most model comparison scenarios, one of the models is white-box accessible (\eg the IP owner's model or a public pretrained model), which should be selected as the target model. A random one is selected if both models are black-box.

We assume there is a set of normal input samples $X_{seed}$ available for the target model $f$, which is reasonable since the model's prediction APIs are available and the functionalities are usually known.

Directly using the normal inputs to extract the decision logic of the models is problematic for our purpose, since the normal inputs can only trigger normal knowledge that may be shared by unrelated models.
For example, suppose $f$ and $g$ are two irrelevant image classifiers trained from scratch (\ie there is no knowledge reuse between them), $x_{cat1}$ and $x_{cat2}$ are two normal images of cat and $x_{dog}$ is a normal image of dog, then it is highly possible that:
$$\dist(f(x_{cat1}),f(x_{cat2})) \approx \dist(g(x_{cat1}), g(x_{cat2})) \approx 0$$
$$\dist(f(x_{cat1}),f(x_{dog})) \approx \dist(g(x_{cat1}), g(x_{dog})) > 0$$
which means that the reaction patterns of $f$ and $g$ on the normal inputs may be indistinguishable.
Such similarity between the decision patterns is caused by the intrinsic features in the normal inputs and the commonly-agreed labels of them - such knowledge of normal inputs is implied in the datasets and is obtained by different (unrelated) models trained on the similar datasets.

Thus, to achieve our goal (model reuse detection), we ought to generate test inputs that can trigger the model-specific knowledge that is shared by models with knowledge reuse while not shared by any other unrelated models.

Inspired by prior work on adversarial attacks \cite{szegedy2013intriguing,demontis2019why_transfer,rezaei2019target_agnostic_attack} that discovered the imperfect decision boundaries are one of the main reasons for adversarial vulnerability, we attempt to address the input generation problem from the decision boundary perspective.

We argue that the decision boundaries of models with reused knowledge are similar. For example, in transfer learning, the decision boundary of the teacher model is copied into the student and fine-tuned on the student dataset. The fine-tuning will not alter the decision boundary significantly, instead, it only adjusts the decision boundary to fit the student dataset. Similarly, other model reuse methods like pruning and quantization are also designed to inherit the decision boundary rather than changing it.
Thus, if we can precisely interpret the decision boundaries with the test inputs, it will help identify the reusing relation between DNNs.

\begin{figure}
    \centering
    \includegraphics[width=5cm]{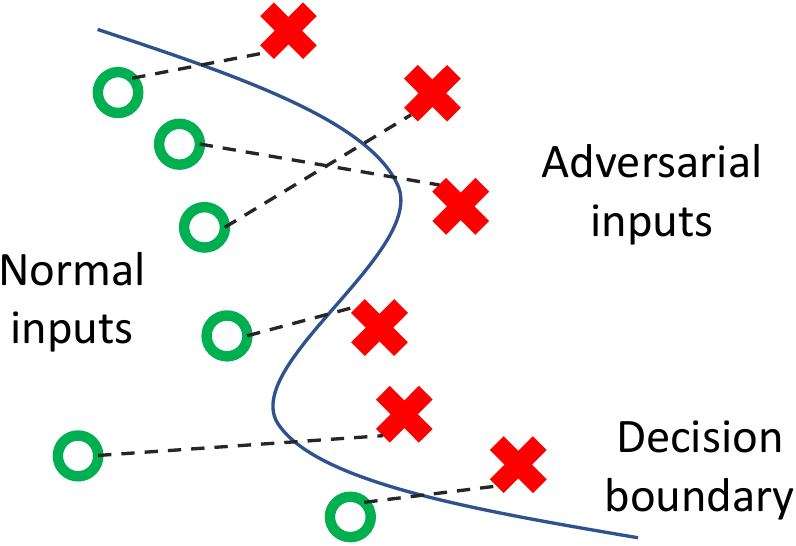}
    \caption{Illustration of the decision boundary depicted by normal and adversarial input pairs.}
    \label{fig:boundary}
\end{figure}

We combine adversarial inputs and normal inputs to create the test inputs in \sys. The intuition behind this idea is shown in Figure~\ref{fig:boundary}.
Specifically, for each normal input $x_i \in X$, we generate a corresponding adversarial input $x_i'$ by adding small perturbation to $x_i$. The normal input $x_i$ usually lies inside the decision boundary, and the output produced by the normal input typically reflect the general knowledge shared by similar but unrelated models. The adversarial input $x_i'$, on the other hand, lies around the decision boundary and the corresponding output is mainly determined by the model-specific imperfect decision boundary. By using each other as a reference, the decision distance between normal and adversarial inputs can convey the knowledge exclusively shared between reused models, \ie
$$\dist(f(x_{i}),f(x_i')) \approx \dist(g(x_{i}),g(x_{i}')),\ if\ f \sim g$$
$$\dist(f(x_{i}),f(x_i')) \neq \dist(g(x_{i}),g(x_{i}')),\ if\ not\ f \sim g$$

To generate the adversarial inputs $X' = \{x_i', x_2', ...\}$ from the normal inputs $X = \{x_i, x_2, ...\}$, we introduce two criteria to measure the quality of generated test inputs, including \textit{intra-input distance} that represents the element-wise distance between the outputs produced by $X$ and $X'$ and \textit{inter-input diversity} that represents the diversity of outputs produced by $X'$.
$$\divergence_f(X', X) = \mean_{i=0,1,...,|X|}\big\{||f(x_i') - f(x_i)||_2\big\}$$
% and \textit{inter-input diversity} that represents the diversity of outputs produced by $X'$:
$$\diversity_f(X') = \mean_{x'_i,x'_j \in X'}\big\{||f(x'_i) - f(x'_j)||_2\big\}$$
The implication is two-fold: First, $\divergence_f(X', X)$ implies the strength of the adversarial inputs, \ie the decision boundary depicted by $X$ and $X'$ are more transferable if $\divergence$ is larger. Second, $\diversity_f(X')$ indicates the coverage of behaviors produced by the inputs. There are more standard neuron coverage metrics proposed by prior work \cite{pei2017deepxplore,ma2018deepgauge}, but we use the output-based criteria $\diversity_f(X')$ since we don't assume the access to the model internal structure.
The quality score of a set of adversarial inputs $X'$ is measured by:
\begin{equation}
\label{equation:input_score}
score(X') = \divergence_f(X', X) + \lambda \: \diversity_{f}(X')
\end{equation}
where $\lambda$ is a hyperparameter to balance the two criteria.

The goal of input generation is to find:
\begin{equation}
\label{equation:input_goal}
\hat{X'} = \argmax_{X'} \: score(X')
\end{equation}

There are several ways to solve Equation~\ref{equation:input_goal}. If $f$ is white-box accessible, the adversarial inputs can be generated through gradient ascent \cite{madry2017towards,szegedy2013intriguing}. Specifically, we select a target output $f(X')$ that maximizes Equation~\ref{equation:input_goal} and use the PGD attack \cite{madry2017towards} to generate $X'$ that can minimize the loss between $f(X')$ and $f(X)$.
% Specifically, we select a target output $Y'$ that 

In cases where the target model is also black-box, we introduce a criteria-guided search algorithm (as shown in Algorithm~\ref{algorithm:black-box-input-gen}) to generate test inputs for the model $f$ by gradually mutating the seed inputs towards to the goal.
The algorithm is inspired by prior work on mutation testing \cite{jia2010mutation} and black-box adversarial attack \cite{guo2019simple,demir2019deepsmartfuzzer} while tailored for our objective in terms of mutation index selection and mutation operation.

\begin{algorithm}[t]
\caption{Black-box input generation in \sys.}\label{algorithm:black-box-input-gen}
%   \scriptsize
\DontPrintSemicolon
% \SetAlgoNoEnd
% \LinesNumbered
\KwIn{$f$: the target model, $X$: the set of seed inputs, $\lambda$, $\epsilon$, $N$: hyperparameters to control divergence-diversity balance, mutation strength, and number of iterations.\\}
% \Begin{
\nl {initialize inputs $X' \leftarrow X$} \\
\nl initialize $score \leftarrow \divergence_f(X, X') + \lambda \: \diversity_{f}(X')$ \\
\nl \For{$i$ from 1 to $N$}{
\nl {compute $\divergence_f(X, X')$ and $\diversity_f(X')$}\\
\nl {$indices \leftarrow I_{low\_divergence} \: \bigcup \: I_{low\_diversity}$}\\
\nl {$pos \leftarrow random\_pick(X[0].shape)$}\\
\nl {compute $X'_{left}$ by adding $-\epsilon$ to $X'[indices][pos]$}\\
\nl {compute $X'_{right}$ by adding $\epsilon$ to $X'[indices][pos]$}\\
\nl {compute $score_{left}$ and $score_{right}$ using $X'_{left}$ and $X'_{right}$}\\
\nl \uIf{$score_{left} > score$ and $score_{left} > score_{right}$}{
\nl {$X' \leftarrow X'_{left}$, $score \leftarrow score_{left}$}
}
\nl \ElseIf{$score_{right} > score$}{
\nl {$X' \leftarrow X'_{right}$, $score \leftarrow score_{right}$}
}
}
\nl \KwRet{the generated test inputs $X'$}
\end{algorithm}

Given a target model $f$ and a set of seed inputs $X$, we generate the test inputs through $N$ mutating iterations. In each iteration, we select a subset of input samples (named mutation inputs) in $X$ that are the primary cause of low divergence and low diversity (line 6). $I_{low\_divergence}$ are the indices of inputs where each $j \in I_{low\_divergence}$ satisfies $||f(x'_j) - f(x_j)||_2 < \divergence_f(X, X')$, \ie the divergence between $x_j$ and $x'_j$ is lower than the average thus $x'_j$ should be mutated.
To compute $I_{low\_diversity}$, we first calculate the distance $||f(x'_j) - f(x'_k)||_2$ between each input pair $x_k, x_l \in X'$. The input pairs with smaller distances are more responsible for the low diversity thus should be mutated. We set $I_{low\_diversity}$ to the indices of the first $r \times n$ input pairs, where $r$ is a hyperparameter to control the size of $I_{low\_diversity}$. $r$ is set to 0.5 by default so that $I_{low\_diversity}$ will contain no more than $n/2$ indices.

After selecting the mutation inputs, we randomly pick a position $pos$ in the input shape (\eg a pixel in the input image) to perform the mutation operation. We obtain two sets of inputs $X'_{left}$ and $X'_{right}$ that are generated by adding a small perturbation $\epsilon$ to or subtracting $\epsilon$ from the mutation position $pos$ in each mutation input.
We compute the scores for $X'_{left}$ and $X'_{right}$ respectively with Equation~\ref{equation:input_score} and update the test input set $X'$ if the score is improved.
The mutation process is repeated for $N$ iterations and the final input set $X'$ is produced as the result of input generation.

% If $f$ is a classification model, solving Equation~\ref{equation:input_goal} can be simplified. First, run an inference pass to find the outputs of $f$ on normal inputs $X$, then decide the target labels of $f(X')$ by maximizing the criterion defined in Equation~\ref{equation:input_score}, and the inputs can be generated with state-of-the-art white-box \cite{madry2017towards} and black-box \cite{guo2019simple} targeted adversarial attack methods.

\subsection{Similarity Comparison}
\label{sec:approach:decision_logic_analysis}

With the normal seed inputs $X$ and adversarial inputs $X'$ generated in Section~\ref{sec:approach:test_input_generation}, we are able to compute the decision distance vectors (DDVs) for the models under comparison.

First, the normal inputs $X$ and adversarial inputs $X'$ are combined to form a list of input pairs $\XP = \{(x_1,x_1'), (x_2,x_2'), ..., (x_n, x_n')\}$, where $x_i \in X$, $x_i' \in X'$, and $n$ is the number of inputs in $X$.
The decision distance vector (DDV) is defined as:
\begin{definition}
\textbf{(Decision distance vector)} Given a list of input pairs $\XP$, the decision distance vector (DDV) of a model $f$ is a float vector $\DDV_f(\XP) = <v_1, v_2, ..., v_n>$, in which each element $v_i = \dist(f(x_i), f(x_i'))$ is the distance between the responses of $f$ produced by $x_i$ and $x_i'$.
\end{definition}

For each input pair, DDV measures the distance between the outputs produced by the two inputs. Since the outputs are produced by the same model, the outputs $f(x_i)$ and $f(x_i')$ are comparable.
The distance metric $\dist$ is the Cosine distance here if $f(x_i)$ is a 1-D array (\eg when $f$ is a classifier), since Cosine distance is good at comparing different scales of vectors.

A DDV basically captures the decision pattern of a model on the test inputs.
If two models are similar, they would have similar patterns when measuring the distance between each pair of test inputs.
The concept is analogous to testing two people with the same quiz questions, they would give similar answers if they have common knowledge.

The length of DDV equals to the number of input pairs in $\XP$ used to compute the DDV.
By using the same set of profiling input pairs to compute the DDVs for different models, we are able to generate DDVs with fixed length and common semantics. Thus the DDVs are comparable across different models, and the model similarity can be measured through DDV comparison. Specifically,
$$ \similarity(f, g) = Cosine\_similarity(\DDV_f, \DDV_g)$$

\subsection{Threshold to Identify Model Reuse}

The similarity score computed by \sys is an indicator of how likely a model is reused from another. However, in practice it is usually desirable to have a threshold to decide whether it is a reuse. Defining a global threshold is difficult because the range of similarity scores may differ across various model types. Instead, we opt for a data-driven model-specific threshold, \ie when we want to determine whether a suspect model $g$ is a reuse of the target model $f$, we first collect (or generate) several reference models that are similar to $f$ but not built upon $f$ (\eg models trained with the same dataset of $f$ from scratch or built upon other pretrained models). Then the threshold can be determined as the maximum of the similarity scores obtained by the reference models. We will show in Section~\ref{section:eval:performance} that such threshold is feasible and effective.

%% file: sec_evaluation.tex
\section{Evaluation}

Our evaluation aims to address the following research questions:

\begin{enumerate}
    \item What is the performance of \sys? Is it able to correctly detect different types of model reuses? (\S\ref{section:eval:performance})
    \item How effective are the inputs generated with mutations in the complete black-box setting? (\S\ref{section:eval:blackbox})
    \item How do different configurations of \sys affect the similarity comparison performance? (\S\ref{section:eval:configuration})
    \item Can \sys be applied to real-world deep learning apps to analyze model similarity? (\S\ref{section:eval:real_world})
\end{enumerate}

\subsection{Experiment Setup}

\textbf{The \bench Benchmark}.
To evaluate our method, we create a benchmark named \bench for model similarity comparison.

We use state-of-the-art image classification models and datasets that commonly appear in transfer learning literature to construct the benchmark. The source models to transfer knowledge from are ResNet18 \cite{he2016deep} and MobileNetV2 \cite{sandler2018mobilenetv2} pretrained on ImageNet \cite{deng2009imagenet}, and the datasets to transfer knowledge into are Oxford Flowers 102 (Flower102 for short) \cite{Nilsback08} and Stanford Dogs 120 (Dog120 for short) \cite{KhoslaYaoJayadevaprakashFeiFei_FGVC2011}.
The other models are generated from the base models using different model reuse methods with varying configurations.

% We only use CNN models in \bench because most existing model reuse methods are implemented on CNNs. Our method should support other types of models (e.g. RNNs) since it doesn't rely on CNN-specific features. We leave adding other types of model to \bench to future work.

\begin{table*}[]
\centering
\caption{The \benchN models included in \bench benchmark. The pretrained models are downloaded from the Internet. The transferred models are built upon the pretrained models. The pruned/quantized/distilled/stolen models are based on the transferred models. The retrained models are built from scratch. The ``\#'' column is the number of models trained with the corresponding method and configuration, and the ``Examples'' column shows the names of some models in the category.}
\label{table:eval:modelreuse}
\resizebox{\textwidth}{!}{
\begin{tabular}{p{1.8cm}lp{0.4cm}p{6.8cm}l}
\toprule
\textbf{Method} & \textbf{Configuration} & \textbf{\#} & \textbf{How to generate} & \textbf{Examples}    \\
\midrule
\textbf{Pre-training} & - & 2 & \pbox{6.8cm}{- Train ResNet18 and MobileNetV2 on ImageNet dataset.} & \pbox{7cm}{\texttt{train(ResNet)}\\ \texttt{train(MbNet)}} \\ \midrule
\textbf{Transfer learning} & Tune 10\% layer & 4 & - Transfer each source model to each target dataset (Flower102 and Dog120), fine-tune the last 10\% layers. & \texttt{train(ResNet)-transfer(Flower102,0.1)} \\
& Tune 50\% layers & 4 & - Transfer each source model to each target dataset, fine-tune the last 50\% layers. & \texttt{train(MbNet)-transfer(Dog120,0.5)} \\
& Tune all layers & 4 & - Transfer each source model to each target dataset, fine-tune all 100\% layers. & \texttt{train(MbNet)-transfer(Dog120,1.0)}    \\
\midrule
\textbf{Pruning} & Prune ratio 0.2 & 12 & - Prune 20\% weights in each transferred model and fine-tune. & \texttt{train(ResNet)-transfer(Dog120,0.5)-prune(0.2)}         \\
& Prune ratio 0.5 & 12 & - Prune 50\% weights in each transferred model and fine-tune. &                                      \texttt{train(MbNet)-transfer(Flower102,0.1)-prune(0.5)}  \\
& Prune ratio 0.8 & 12 & - Prune 80\% weights in each transferred model and fine-tune. & \texttt{train(ResNet)-transfer(Dog120,1.0)-prune(0.8)}                                   \\
\midrule
\textbf{Quantization} & INT8 & 12 & - Compress each transferred model using post-training weight quantization. &              \texttt{train(ResNet)-transfer(Flower102,0.1)-quant}                \\
\midrule
\textbf{Knowledge distillation} & same arch & 12 & - Distill each transferred model to a target model with the same architecture using feature distillation. & \texttt{train(MbNet)-transfer(Dog120,0.5)-distill}  \\
\midrule
% \textbf{Stealing} & same arch & 12 & - Use the output of each transferred model to train a target model with the same architecture. & \texttt{train(ResNet)-transfer(Dog120,1.0)-steal(ResNet)}   \\
% & different arch & 12 & - Use the output of each transferred model to train a target model with different architecture. & \texttt{train(ResNet)-transfer(Flower102,0.5)-steal(MbNet)}  \\
\textbf{Stealing} & different arch & 12 & - Use the output of each transferred model to train a target model with different architecture. & \texttt{train(ResNet)-transfer(Flower102,0.5)-steal(MbNet)}  \\
\midrule
\textbf{Retraining} & - & 28 & \pbox{6.8cm}{- Train ResNet18 and MobileNetV2 on each target dataset from scratch. Use model reuse techniques to generate more variations.} & \pbox{7cm}{\texttt{retrain(ResNet)}\\ \texttt{retrain(MbNet)-transfer(Flower102,0.5)}} \\
\bottomrule
\end{tabular}
}
\end{table*}

The complete list of models included in \bench is shown in Table~\ref{table:eval:modelreuse}. In total we have \benchN models, including 2 pretrained source models, 84 student models (12 transferred + 36 pruned + 12 quantized + 12 distilled + 12 stolen), and 28 retrained models. Each of the 84 student models is built from one of the two pretrained source models, and the 28 retrained models are trained from scratch.

Based on the models, we obtain \benchNP pairs of similar models (\ie model pairs in which one model reuses the knowledge of another model, and should be detected as similarity), including 84 direct-reuse model pairs and 60 combined-reuse model pairs.
Each direct-reuse model pair is a student model with its direct teacher model. Each of the 60 combined-reuse model pairs is generated with a combination of two reuse methods (transfer learning + model compression) from the corresponding source model.
Such combined reuse is common in real-world deep learning applications where both the task and the model size are customized.

\textbf{Baselines}.
To our best knowledge, there is no existing work aimed to address the same problem as ours. However, there are similar concepts discussed in related fields such as transfer learning, watermarking, etc. Thus we implement several baselines for comparison:
\begin{enumerate}
    \item \textit{WeightCompare} measures the model similarity directly based on weight comparison. Specifically, the similarity between model $f$ and $g$ is calculated by $\frac{\#\text{identical layers between f and g}}{min(\#\text{layers of f},\ \#\text{layers of g})}$, where two layers are identical if and only if their structures and weights are the same.
    \item \textit{FeatureCompare} compares the feature maps produced by the same set of $N$ normal inputs. Suppose $f^{feat}(x)$ is the feature map of the last Conv layer in model $f$ produced by input $x$, then the model similarity between $f$ and $g$ is calculated by $mean_{i=1}^N\{cosine(f^{feat}(x_i), g^{feat}(x_i))\}$.
    \item \textit{Fingerprinting} computes a fingerprint of the teacher model and check the fingerprint against other models to measure similarity. The idea \cite{cao2019ipguard,lukas2019dnn_fingerprinting} is to fingerprint a model $f$ with a set of adversarial inputs $X$ and their predicted label $Y_f$. Given a new model $g$, the IP ownership is verified by checking whether $Y_g \approx Y_f$. We use the same inputs as ours to compute fingerprints and calculate model similarity as $sim(Y_f, Y_g)$.
\end{enumerate}
\textit{WeightCompare} and \textit{FeatureCompare} are while-box methods since they require reading the weights or feature maps. \textit{Fingerprinting} is a black-box method like ours. We also considered other baselines such as directly comparing the model outputs (\textit{OutputCompare}), but since \textit{Fingerprinting} is also based on output comparison, it should be able to represent the performance of \textit{OutputCompare}.

\textbf{Implementation and Test Environment}. \sys was implemented with PyTorch 1.3 and Tensorflow 2.0 using Python 3.6. Unless otherwise noted, we assume the target model is white-box accessible and the suspect model is black-box, and the test inputs are generated using gradient ascent. The number of test inputs was set to 100 and the hyperparameters $\lambda$, $\epsilon$ and $N$ were set to 0.5, 0.06, and 20,000 by default in our implementation. The benchmark dataset was generated on a GPU cluster, and the experiments were conducted on a Linux Server with 2 Intel Xeon CPUs and 2 GeForce GTX 1080Ti GPUs. It takes around 18 seconds for \sys to compare a pair of models.

\subsection{Correctness on \bench Benchmark}
\label{section:eval:performance}

\begin{table*}[]
\centering
\caption{The similarity comparison result of \sys and other baselines on \bench benchmark. Feas. and corr. are the abbreviations of feasibility and correctness respectively.}
\label{table:eval:reuse_detect}
\resizebox{\textwidth}{!}{
\begin{tabular}{c|c|c|cr|cr|cr|cr}
\toprule
\multicolumn{2}{c|}{\multirow{2}{*}{Reuse method}} & \multirow{2}{*}{\#Models} & \multicolumn{2}{c|}{WeightCompare} & \multicolumn{2}{c|}{FeatureCompare} & \multicolumn{2}{c|}{Fingerprinting} & \multicolumn{2}{c}{ModelDiff (ours)} \\
\multicolumn{2}{c|}{} & & Feas. & Corr. & Feas. & Corr. & Feas. & Corr. & Feas. & Corr. \\
\midrule
\multirow{9}{*}{Direct reuse} 
& Transfer - tune 10\%      & 4     & \cmark & 100\% & \cmark & 100\% & \xmark & - & \cmark & 100\% \\
& Transfer - tune 50\%      & 4     & \cmark & 100\% & \cmark & 100\% & \xmark & - & \cmark & 100\% \\
& Transfer - tune 100\%     & 4     & \cmark & 0\% & \cmark & 100\% & \xmark & - & \cmark & 100\% \\
& Prune 20\%                & 12    & \cmark & 0\% & \cmark & 100\% & \cmark & 100\% & \cmark & 100\% \\
& Prune 50\%                & 12    & \cmark & 0\% & \cmark & 100\% & \cmark & 100\% & \cmark & 100\% \\
& Prune 80\%                & 12    & \cmark & 0\% & \cmark & 100\% & \cmark & 66.7\% & \cmark & 100\% \\
& Quantize                  & 12    & \cmark & 100\% & \cmark & 100\% & \cmark & 100\% & \cmark & 100\% \\
& Distill - same arch       & 12    & \cmark & 0\% & \cmark & 50.0\% & \cmark & 75.0\% & \cmark & 100\% \\
% & Steal - same arch         & 12    & \cmark & 0\% & \cmark & 0\% & \cmark & 0.0\% & \cmark & 50.0\% \\
& Steal - different arch    & 12    & \xmark & - & \xmark & - & \cmark & 0\% & \cmark & 0\% \\
\midrule
\multirow{3}{*}{Combined reuse}
& Transfer + prune          & 36    & \cmark & 0\% & \cmark & 100\% & \xmark & - & \cmark & 100\% \\
& Transfer + quantize       & 12    & \cmark & 66.7\% & \cmark & 100\% & \xmark & - & \cmark & 100\% \\
& Transfer + distill        & 12    & \cmark & 0\% & \cmark & 50.0\% & \xmark & - & \cmark & 100\% \\
\midrule
\multicolumn{2}{c|}{Overall} & 144 & 91.7\% & 21.2\% & 91.7\% & 90.9\% & 50.0\% & 73.6\% & 100\% & 91.7\% \\
\bottomrule
\end{tabular}
}
\end{table*}

We first ran \sys and the baseline methods on the \bench benchmark to test their performance of similarity comparison.
% We considered both direct reuses (where the suspect model is built from the teacher model using only one model reuse method) and combined reuses (where the suspect model is built from the teacher model with a transfer-and-compress combination, which is common in mobile model development).
% We considered both direct reuses and combined reuses, and the detailed results are shown in Table~\ref{table:eval:reuse_detect}.

For each of the \benchNP reused model pairs in \bench, we generate reference model pairs by randomly replacing one of the two models with an unrelated one (\eg a model with different source model or a retrained model). Thus each reused model pair has 71 reference model pairs.
% \yuanchun{paused here} For each student model, the comparator is invoked to compare the model with its corresponding reused model and 5 randomly-chosen dissimilar models. The following metrics are measured:
When evaluating a model comparison method, each reused model pair and its corresponding reference model pairs are fed into the comparator, and the following two metrics are computed for each method:
\textbf{Feasibility}. Whether the comparator can be used to compare the reused model pair.
\textbf{Correctness}. Whether the comparator can distinguish the reused model pair from reference model pairs (\ie whether the similarity score of the reused model pair is higher than all reference model pairs).

% The model pairs can be used to evaluate a given model similarity comparator. The comparator should compute a similarity score for each pair of models. 
% The 156 pairs of similar models are labeled as \emph{positive} model pairs, and for each \emph{positive} model pair, we generate 5 \emph{negative} model pairs by randomly replacing one of the two models with a dissimilar one (\eg the models with different source model or the retrained models).
% % \yuanchun{paused here} For each student model, the comparator is invoked to compare the model with its corresponding reused model and 5 randomly-chosen dissimilar models. The following metrics are measured:
% When evaluating a model comparator, each \emph{positive} model pair and the five corresponding \emph{negative} model pairs are fed into the comparator, and the following two metrics are computed:
% \textbf{Feasibility}. Whether the similarity comparator can be applied to compare the models.
% \textbf{Correctness}. Whether the comparator can distinguish the \emph{positive} model pair from 5 other \emph{negative} model pairs (\ie whether the similarity score computed for the \emph{positive} model pair is higher than other model pairs).
% % \item \textbf{Confidence}. The gap between the score computed for the reused model and the maximum score for other models.

The results are shown in Table~\ref{table:eval:reuse_detect}.
First of all, \sys achieved 100\% feasibility, meaning that \sys can measure the similarity between all types of models, including those with different model architectures or output spaces, while all other baselines are not 100\% feasible.
Specifically, the white-box approaches are unable to process models with different architecture, which is a disadvantage since the cross-structure distillation techniques \cite{hinton2015distilling} are gaining popularity. \textit{Fingerprinting} is not designed for models with different underlying tasks, thus was unable to detect any reuse related to transfer learning.

% Regarding correctness, 
\sys achieved overall correctness of \correctness, outperforming all the other baseline methods including the white-box approach \textit{FeatureCompare}. Specifically, \sys was able to precisely identify the reused models generated with all model reuse methods except for stealing. \textit{FeatureCompare} is also precise on most normal reuses.
However, the qualitative differences between \textit{FeatureCompare} and \sys are notable.
First, \textit{FeatureCompare} is a white-box approach because it requires access to the intermediate feature of the compared models.
Second, \textit{FeatureCompare} assumes that the compared models have a common feature layer for comparison. However, finding the common layer between two models is non-trivial or even impossible, especially if the suspect model is generated through knowledge distillation or modified purposely. 
Table~\ref{table:eval:reuse_detect} has showed that \textit{FeatureCompare} was not as effective on models generated with knowledge distillation that may lead to significant internal feature change by retraining the weights from scratch.

The stolen models were difficult to detect with all methods. Stealing a model is almost equivalent to retraining it, and the teacher model is only used to generate a training dataset.
% Nevertheless, \sys was able to achieve the highest correctness on stolen models compared with baselines.
How to identify models generated with stealing remains a challenging problem.

\begin{figure}
    \centering
    \includegraphics[width=7.2cm]{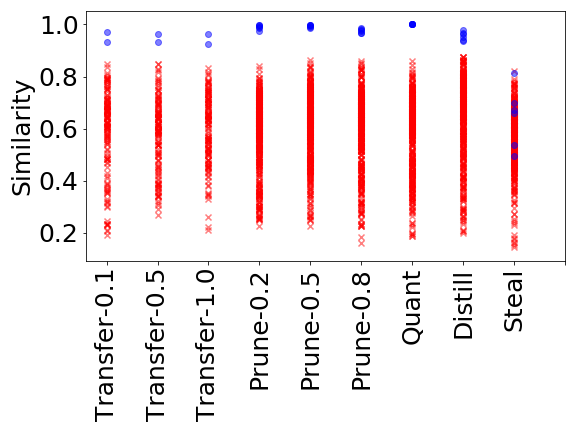}
    \caption{The similarity score distribution of different reuse methods. Blue dots and red crosses are the similarity scores obtained by reused models and irrelevant models respectively.}
    \label{fig:eval:similarity_dist}
\end{figure}

Figure~\ref{fig:eval:similarity_dist} shows the distribution of the similarity scores computed by \sys on the ResNet-based models. In most model reuse cases that \sys can correctly detect, we observe a clear gap between the scores achieved by the reused model pairs and the reference model pairs, meaning it's easy to identify reused models with a threshold. The gap is smaller when the models are generated with knowledge distillation, which is intuitive since distillation would reset the model parameters rather than reuse the parameters from the teacher, thus less decision boundaries are inherited.

% \begin{figure}[]
%     \centering
%     \includegraphics[width=8.4cm]{figure/similarity_with_ratio.png}
%     \caption{Similarity scores of models generated with different tune ratios or prune ratios.}
%     \label{fig:eval:similarity_with_ratio}
% \end{figure}

% Figure~\ref{fig:eval:similarity_with_ratio} shows the impact of different transfer ratios (\ie fine-tuning 10\%, 50\%,, 100\% parameters during transfer learning) and prune ratios (\ie pruning 20\%, 50\%, and 80\% parameters during model compression) on the similarity. The similarity decreases as the increase of the ratio, which is intuitional as more differences are introduced between the teacher model and the student model by tuning or pruning more parameters of the student.
% % However, the similarity remains high (>90\%) even if the ratios are close to 1.0.

\begin{figure}[]
\centering
\includegraphics[width=6.6cm]{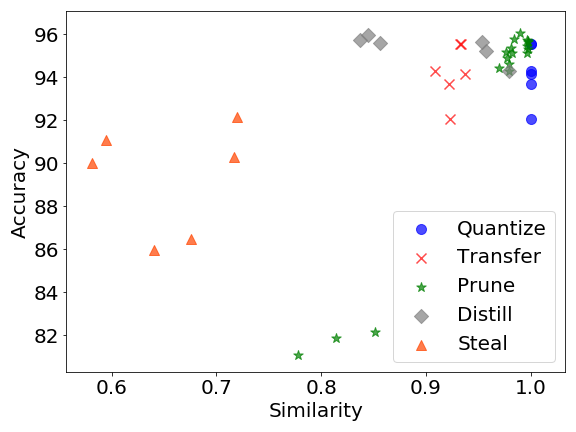}
\caption{Relation between model similarity and test accuracy on Flower102 dataset.}
% \label{fig:eval:sim_acc}
\label{fig:eval:sim_acc}
\end{figure}

% To interpret the similarity score computed by \sys, we visualized the relation of the measured similarity to the test accuracy achieved by the student model and vulnerability transferability (the error rate of the student model on the adversarial inputs generated with the teacher model), as shown in Figure~\ref{fig:eval:similarity_explain}. We notice that both the model accuracy and the vulnerability transferability are proportional to the similarity, \ie the higher similarity between the teacher and the student means more useful knowledge is transferred to the student, while more defects are inherited as well.

% However, the proportional relation is not strictly true. For example, the models generated with transfer learning with different fine-tune ratios had achieved similar accuracy, while their similarities with the teacher model are different (range from 0.7 to 1). This implies the possibility of partially reusing the knowledge of the teacher model, \ie reusing the good knowledge to achieve high accuracy while avoiding being too similar with the teacher, so that the vulnerability propagation can be mitigated.

To further interpret the similarity scores, we visualized the relation between each student model's similarity score and its test accuracy in Figure~\ref{fig:eval:sim_acc}. We noticed that the student models with higher accuracy typically have higher similarity scores, because more useful knowledge is transferred from the teacher. The test accuracy of the models generated with stealing attack is lower, meaning that they didn't reuse much useful knowledge although they are harder to detect. 
Surprisingly, although some models (pruning 80\% weights in MobileNet) didn't inherit much useful knowledge from the teacher (thus had a poor accuracy), we are still able to correctly detect them.

% \begin{figure}[]
% \centering
% \includegraphics[width=0.8\linewidth]{figure/sim_adv_Flower102.pdf}
% \caption{Relation between model similarity and vulnerability transferability.}
% \label{fig:eval:sim_dir}
% \end{figure}

% We also analyzed the relation between the measured similarity with vulnerability transferability (the error rate of the student model on the adversarial inputs generated with the teacher model), as shown in Figure~\ref{fig:eval:sim_dir}. We notice that the vulnerability transferability is also roughly proportional to the similarity, \ie the higher similarity between the teacher and the student means more defects are inherited.

\subsection{Complete Black-box Setting}
\label{section:eval:blackbox}

\begin{figure}[]
    \centering
    \includegraphics[width=7cm]{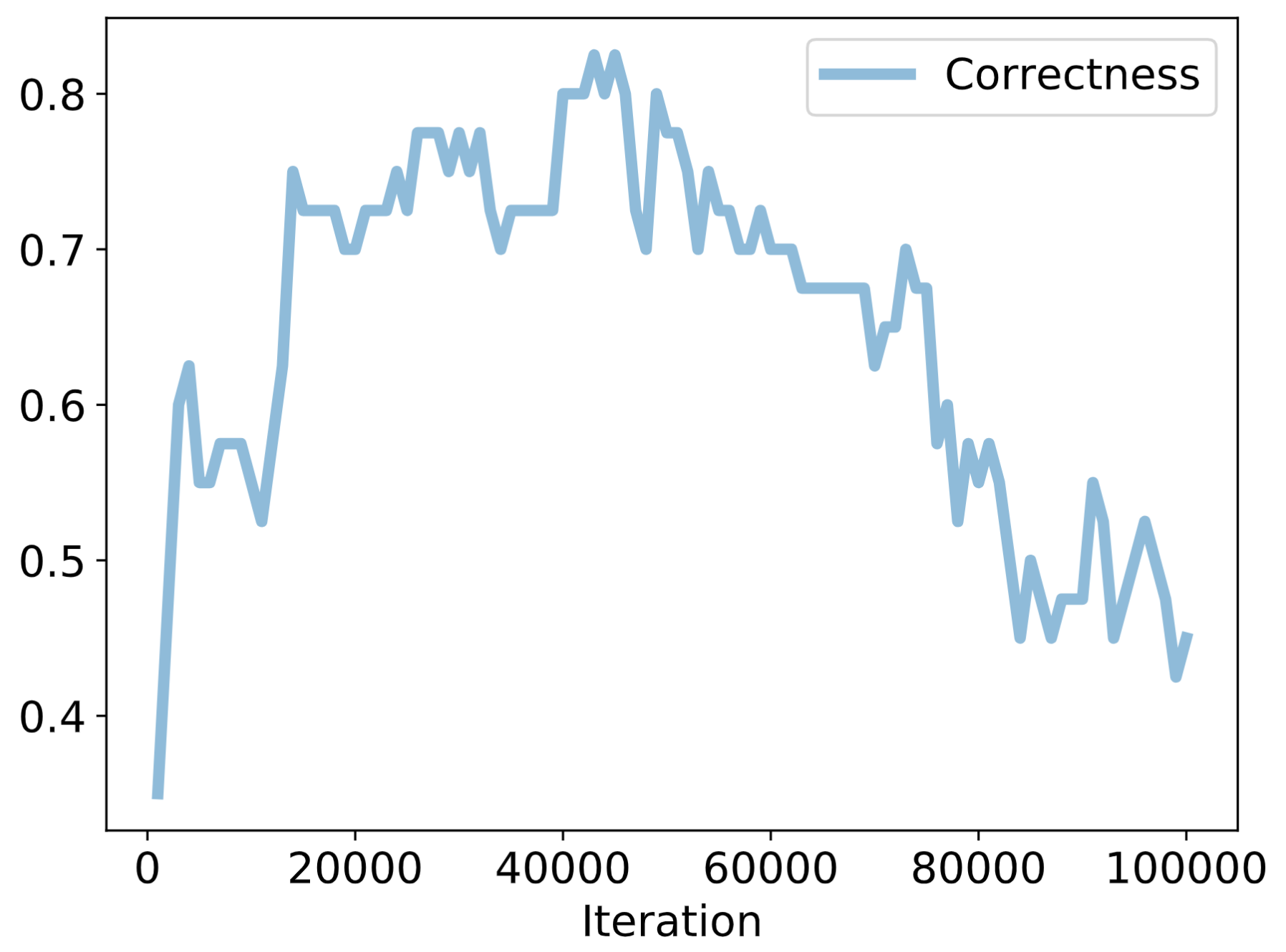}
    \caption{Progressive correctness achieved with different numbers of input mutations in complete black-box setting.}
    \label{fig:eval:progressive_correctness}
\end{figure}

% The computation time of our method is mainly determined by how many mutation iterations are needed for input generation.
In complete black-box settings, both models under comparison are black-box. Thus the test inputs can only be generated with mutation (Algorithm~\ref{algorithm:black-box-input-gen}) rather than gradient ascent.
To evaluate the performance of \sys in this setting, we used the two pretrained models (ResNet18 and MobileNetV2) to generate test inputs, and measured the correctness achieved by the generated inputs every 1,000 iterations.
The result is shown in Figure~\ref{fig:eval:progressive_correctness}. 

% Overall, the performance achieved with inputs generated with mutations were less stable.
At first, the correctness improved quickly as more mutations were performed, since the generated inputs became better at depicting the decision boundaries.
The correctness went to 70\%-80\% with roughly 20,000 to 60,000 mutations (which took around 1-3 hours), meaning that \sys was able to measure model similarity with a reasonable accuracy in complete black-box settings.
However, the correctness started to drop when the number of mutations was larger, because too many mutations had made the adversarial inputs more transferable to other irrelevant models.
In practice, we should limit the number of mutations to avoid such issues.
We leave more stable and effective black-box generation of adversarial inputs that are only transferable to reused models as future work.

% The result shows that \sys can achieve a 100\% correctness in less than 10,000 mutation iterations (which cost roughly 10 minutes). The input quality score continues increasing and the similarity score measured for reused model pairs slowly converges by adding more iterations. Since the result is based on popular large-scale models, we believe the performance overhead of \sys is acceptable in practice.

% \subsection{Interpretation of the similarity score}
% \label{section:eval:interpretation}

% \begin{figure}[]
% \centering
% \includegraphics[width=0.8\linewidth]{figure/sim_acc_Flower102.pdf}
% \caption{Relation between model similarity and test accuracy.}
% % \label{fig:eval:sim_acc}
% \label{fig:eval:similarity_explain}
% \end{figure}

\subsection{Configuration Analysis}
\label{section:eval:configuration}

\begin{table}[]
\centering
\caption{The relative correctness of \sys under different configurations as compared to the default setting.}
\label{table:eval:relative}
\begin{tabular}{cc}
\toprule
Variation         & Relative correctness \\
\midrule
Random noise as seed inputs      & 0.59                                \\
Less (10) or more (200) seed inputs  & 0.82, 1.00                                \\
Irrelevant images as seed inputs & 1.00                                \\
All normal inputs       & 0.61                                \\
All adversarial inputs  & 0.86                               \\
Without diversity       & 0.75          \\
\bottomrule
\end{tabular}
\end{table}

Since the test inputs are critical in \sys to precisely measure the knowledge similarity, we further analyzed how different configurations of test inputs may affect the correctness on the \benchNPdirect direct-reuse model pairs. The results are shown in Table~\ref{table:eval:relative}.

The first three rows discuss the choice of seed inputs. The choice of seed inputs is important for \sys since using random noises as the seed inputs or reducing the number of seed inputs would lead to a correctness drop. However, \sys performs well even if the seed inputs are irrelevant images drawn from other datasets. Thus \sys only requires the inputs to comply with the input distribution of the models under comparison.

The following two rows discuss how effective it is to measure model similarity with normal inputs only or adversarial inputs only. Both the two variations were unable to achieve performance comparable to our default configuration (half adversarial inputs and half normal inputs), which demonstrates the effectiveness of \sys in using adversarial inputs and corresponding normal inputs together to interpret the models' decision boundaries.

The last row discusses the usefulness of the \emph{diversity} metric introduced in Section~\ref{sec:approach:test_input_generation}. By removing the diversity of seed inputs and disabling the diversity criterion in Equation~\ref{equation:input_score}, \sys's overall correctness dropped by 25\%. This demonstrates the usefulness of considering test output diversity when generating test inputs.

\subsection{A Study on Real-world Models}
\label{section:eval:real_world}

We further studied whether \sys can be applied to measure similarity for real-world models by testing it on models extracted from real-world Android apps.

% and whether \sys can measure similarities for real-world models. The study involves the following 3 steps: 1) select a public pretrained model as the source model and generate the test inputs with the source model, 2) crawl models from real-world mobile apps, and 3) for each real-world model, compute its similarity with the source model using \sys. The whole process is similar to vulnerability scanning, in which the source model is regarded as a vulnerable software component, and the goal of scanning is to find other applications that use a similar component that probably contains the vulnerability as well.

We selected MobileNet-V2 \cite{sandler2018mobilenetv2} pretrained on ImageNet \cite{deng2009imagenet} as the source model due to its popularity in mobile apps. The pretrained MobileNet-V2 was obtained from Keras (https://keras.io/), which is commonly used by app developers to download pretrained models. The test inputs used by \sys to measure knowledge similarity were generated from the source model.

% By looking into resource files from mobile apps, we can get lots of real-world models.
To obtain real-world models, we crawled 20,000 popular apps from Google Play and looked for DNN models contained in those apps.
We focused on TFLite models since TFLite is the most popular mobile deep learning framework today and \texttt{.tflite} model files are self-contained and suitable for automated analysis.
In the end, we obtained 149 apps that contain at least a TFLite model. By excluding the models whose input shape was different from the source model, we obtained 35 models for comparison.

% \begin{figure}[]
% \centering
% \includegraphics[width=8cm]{figure/app_model_ddv_hist.png}
% \caption{Histogram of the similarities between 35 real-world models and a public pretrained MobileNet-V2 model.}
% \label{fig:eval:real_world_models_hist}
% \end{figure}

% All the 35 models were successfully processed by \sys to measure the knowledge similarity with the source model. The analysis result is shown in Figure~\ref{fig:eval:real_world_models_hist}. There were 11 models whose similarity scores were larger than 0.6 and 24 models whose similarity scores were lower than 0.5.
% % The real-world models are clearly separated by the 0.5-0.6 gap, so we set the similarity 0.6 as the threshold. We first manually check all cases larger than 0.6, and confirm their correctness.
% Among the 11 high-similarity models, there were 8 models whose file names contain ``MobileNet'', meaning that they were probably built upon the source model or a similar MobileNet model.
% Although we were unable to confirm whether the models were built from the source model since the code of training the app was unavailable,
% we believe such results have already demonstrated the feasibility and effectiveness of \sys in measuring knowledge similarity for real-world models.
% % This indicates the effectiveness of our DDV similarity method in detecting real-world model re-uses. The two MobileNet named cases in 0.2-0.3 gap might represent heavily re-trained MobileNet models.

\begin{figure}[]
\centering
\includegraphics[width=5.0cm]{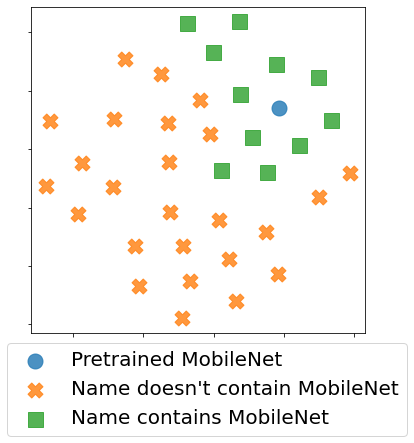}
\caption{t-SNE visualization of the DDVs computed for 35 real-world mobile deep learning models.}
\label{fig:eval:real_world_models_tsne}
\end{figure}

All the 35 models were successfully processed by \sys to compute DDVs with the test inputs. Since there is no ground truth about whether each model is similar with the source model, we were unable to compute the correctness like in Section~\ref{section:eval:performance}. Instead, we grouped the models into two categories based on whether the model name contains ``MobileNet'' and examined whether the DDVs computed for the models with ``MobileNet'' in name are closer to the source model.
% The DDVs computed by \sys can also be used to cluster DNN models.
% Figure~\ref{fig:eval:real_world_models_tsne} presents the visualization of the DDV vectors computed for the models under comparison. We use t-SNE \cite{van2008tsne} to encode DDV vectors into 2-dim points and plot them in the figure. Each point represents a model and is tagged according to whether the model name contains ``MobileNet''.
The result shows that the DDVs of similar models (\eg the MobileNet-related models) are grouped close to each other, which demonstrates the effectiveness of using DDVs to measure model similarity.
Such clustering ability of \sys can potentially be used to analyze model reuse relations at scale in an unsupervised manner.

% In the lower left corner around (-0.5, -0.2), 8 MobileNet-related models are clustered tightly together, which are the 8 reuse cases identified.
% There are also 3 MobileNet related models scattered around (-0.3, -0.1), which are the 3 non-similar cases identified.
% This figure indicates the effectiveness of our DDV vector representation in describing model similarities.

% \yuanchun{TODO add attack transfer experiment.}

%% file: sec_related_work.tex
\section{Related work}

\subsection{Software Similarity Comparison}

Our work is partly inspired by the line of research on code similarity analysis, which has a long history since the emergence of computer software \cite{ragkhitwetsagul2018comparison,rattan2013software_clone_review}. Existing work can be roughly classified into metrics-based, text-based, graph-based, and semantic-based approaches.
The metrics-based approaches \cite{ottenstein1976algorithmic,beighel1984measurement} are mainly focused on computing some metrics from the software and measure the similarity by comparing the metrics.
% For example, Ottenstein \etal \cite{ottenstein1976algorithmic} proposed to use Halstead complexity measures to detect plagiarism in student homework. Berighel \etal \cite{beighel1984measurement} used the numbers of occurrences of particular features as the metrics for comparison. 
Text-based approaches \cite{roy2008nicad,li2006cpminer,schleimer2003moss,yao2018redundancy} view the code snippets as string sequences and compare them using text similarity analysis techniques.
% For example, NICAD \cite{roy2008nicad} proposed to use a Longest Common Subsequence (LCS) algorithm for comparing the text lines of potential clones. CP-Miner \cite{li2006cpminer} and MOSS \cite{schleimer2003moss} are widely-used tools to identify copy-pasted or plagiarized code segments using token-based comparison. These text-based approaches are scalable and efficient since they do not need to understand the code structure, while they may be less effective in finding similarities with syntactic and semantic modifications.
Graph-based approaches parse the programs into a uniform structure (such as abstract syntax tree \cite{baxter1998clone_ast}, control flow graph \cite{chae2013plagiarism_cfg}, program dependence graph \cite{krinke2001similar_pdg}, UI transition graph \cite{li2017droidbot}, etc.) and identify isomorphism between the trees or graphs.
% The main disadvantage of graph-based approaches is the poor scalability since the graph comparison algorithms are mostly NP-complete.
Some recent approaches consider the semantics of code during similarity detection, with the help of advanced NLP and ML techniques \cite{mcmillan2012detecting,xu2017cross_platform_code_sim,liu2018alphadiff_code_sim}.
% These approaches are often used to deal with compiled or obfuscated code due to their great feature extraction abilities.

In cybersecurity research, binary code similarity comparison is attractive due to its rich applications in patch analysis, plagiarism detection, malware detection, and vulnerability search \cite{haq2019survey,ding2019asm2vec,ming2017binsim,xu2017cross_platform_code_sim}.
Jiang~\etal's work~\cite{jiang2009automatic} is the closest to ours, which proposed to compare the final states of two pieces of binary code given the same input. If the pieces of code produce the same output, they are considered equivalent. The same idea is later used in BLEX~\cite{egele2014blanket} and MULTI-MH~\cite{pewny2015cross} for binary similarity detection.
Although the testing-based concept is similar to ours, we deal with DNN models that may serve different tasks.

% Our paper aims to compare the similarity between DNN models rather than the traditional software.
% Most of the existing approaches are inapplicable to model similarity analysis due to the obvious distinction between model and code.
% The work on side-channel attack \cite{nilizadeh2019diffuzz} is conceptually similar to our method since the programs are compared based on their runtime behavior.
% The testing-based similarity analysis method introduced in this paper may be applied to certain types of traditional programs with stochastic or noisy outputs (\eg scheduler, game AI, etc.), although we did not find any such published work.

% Egele~\etal~\cite{egele2014blanket}: Blanket Execution: Dynamic Similarity Testing
% for Program Binaries and Components
% Haq~\etal~\cite{haq2019survey} A survey of binary code similarity: many useful citations

DNN model similarity has also been discussed in the AI and machine learning field \cite{morcos2018insights, raghu2017svcca,kornblith2019similarity}. The main method is canonical correlation analysis (CCA) and the primary purpose is to understand the model internal representation rather than detect reuse. Thus we did not compare with these approaches in this paper.

\subsection{Model Intellectual Property Protection}

To protect the IP right of a model, one way is to avoid model exposure by encrypting it \cite{gilad2016cryptonets}, putting it or part of it into enclaves \cite{tramer2018slalom,zhang2020nnslicer}, etc. Another way is to design mechanisms to enable model IP violation detection.
Here we focus on the detection approaches, including watermarking and fingerprinting.

% Like traditional digital watermarks for images, audios, videos, etc., 
A watermark for a DNN model is usually a marker covertly embedded into the model's weight or output.
% Based on how the ownership is verified, existing approaches can be separated into weight watermarks and output watermarks.
Weight watermarks \cite{uchida2017embedding,chen2019deepattest,chen2019deepmarks} are usually trained into the weights using a parameter regularizer and verified by directly comparing the weights of the intermediate layers. 
% For example, Uchida \etal \cite{uchida2017embedding} proposed to train watermarks into the weights using a parameter regularizer. DeepAttest \cite{chen2019deepattest} used a similar approach while focusing on verifying the watermark in the hardware enclave. 
% These watermarks can be easily removed by overwriting the weights (\eg through a few rounds of fine-tuning).
Output watermarks \cite{adi2018turning,darvish2019deepsigns,fan2019rethinking} are generated by training the DNN model to predict certain outputs (or activation values) on specific inputs, like a backdoor injected into the model.
% The specific inputs that are used for verification can be abstract inputs \cite{adi2018turning}, normal inputs with noise or triggers inserted \cite{zhang2018protecting}, a subset of training inputs \cite{chen2019deepmarks}.
These approaches have in common that the models are overfitted on certain inputs, \ie the watermarks are additional knowledge inserted into the model rather than the intrinsic knowledge of the model. Recent studies have found that the watermarks are not robust against distillation \cite{shafieinejad2019robustness} and retraining \cite{aiken2021neural}.

Unlike watermarking, fingerprinting approaches are focused on post hoc detection of model reuse.
% These approaches typically have two steps, including \textbf{extracting} fingerprints for the models and \textbf{verifying} ownership based on the fingerprints.
For example, IPGuard \cite{cao2019ipguard} is based on the observation that a DNN classifier can be uniquely represented by its classification boundary. Specifically, they find N data points around the classification boundary and use the data points together with the predicted labels as the fingerprint. A suspect model is examined by feeding the N data points and comparing the predicted labels. An IP violation is detected if the suspect model produces outputs similar to the source model.
Lukas \etal \cite{lukas2019dnn_fingerprinting} introduced the concept of conferrable inputs, \ie targeted adversarial inputs that are transferable to surrogate models while not transferable to reference models that are trained independently.
% Similar to IPGuard, Lukas \etal use a set of conferrable inputs as the fingerprint and verify IP by comparing the outputs produced by the inputs.
A major limitation of these approaches is that the outputs of source and suspect model must be in the same label space in order to verify the ownership, which is not true since the suspect models may be transferred to different tasks.

The DDV in this paper can also be viewed as a model fingerprint. However, our method has more broad applicability since we do not require the models to have the same output space.

\subsection{Test Input Generation for DNN}

Prior to our work, DNN testing has been widely discussed \cite{zhang2020mltesting}.
% in the software engineering community.
The primary goal of test input generation for DNNs is to measure or improve the robustness of models to adversarial inputs.
For example, DeepXplore \cite{pei2017deepxplore} introduced the concept of neuron coverage, \ie the ratio of neurons activated by a set of inputs, to describe the adequacy of the input set in revealing the possible behaviors of the model. DeepGauge \cite{ma2018deepgauge} then extended the concept by considering more fine-grained criteria.
Various searching, mutating, and fuzzing techniques \cite{tian2018deeptest,odena2019tensorfuzz,xie2019deephunter} have also been proposed to generate test inputs that can maximize the coverage metrics.

There are also other purposes of DNN testing. For example, 
Ma~\etal~\cite{ma2018mode} proposed to debug model bugs. Zhang~\etal ~\cite{zhang2020whitebox_fairness} and Aggarwal~\etal~\cite{aggarwal2019blackbox_fairness} attempted to test model fairness. Tian~\etal~\cite{tian2019testingbias} are focused on testing the confusion and bias errors in DNNs.
In this work, DNN testing is used for model reuse detection.

% Understanding the decision logic by interpreting the intermediate neurons is also a popular research topic. For example, Carter \etal \cite{carter2019activation} attempted to understand what a DNN has learned by visualizing the neuron activations. Qiu \etal \cite{qiu2019effectivepath} proposed to identify critical neurons by analyzing the contributions of each neuron to the model output. The decision distance vector introduced in this paper is also an abstraction of the decision logic, while it is tailored to capture common knowledge between heterogeneous models.

%% file: sec_limitations.tex
\section{Limitations and Future Work}

\textbf{Other Reuse Methods}.
There are many novel model reuse methods and many variations of existing model reuse methods introduced every day. It is impossible to test all of them, and we only considered the most representative and widely-used ones.

Models generated with other reuse methods may bypass our detection, especially if the student model developers are malicious and aware of our method.
% For example, the effectiveness of our method on detecting models generated through stealing attack is still poor.
How to deal with malicious model reuse methods (\eg model stealing attack) still remains a open problem.
More fundamentally, how to rigorously define DNN knowledge reuse and detect any form of it is an important direction to explore.

\textbf{Models with Different Input Shapes}.
Since our method requires testing the models under comparison with the same set of inputs, it is not able to measure similarity for models with different input shapes. Fortunately, in most model reuse cases, the models with different input shapes are unlikely to be built from each other.
% Model stealing can probably transfer the knowledge to a dataset with a different input shape, which is worth exploring in the future.

% \textbf{Models with unknown input distribution}.
% The seed inputs are important for \sys.

% In the future we will add more representative reuse methods into the \bench benchmark.

\textbf{Other Model Types}.
Currently \sys is only tested on CNN models, while we believe the idea of interpreting a precise and complete decision boundary is general across different types of models. In the future we will try to adapt our method to other types of models such as RNN and Transformers.

\textbf{Distinguishing Teacher and Student}.
\sys currently does not distinguish the direction of model reuse, \ie we are unable to know which is the teacher model and which is the student given two models with knowledge similarity. The ability to detect the reuse direction would be important when one needs to decide the IP ownership among similar models.

%% file: sec_conclusion.tex
\section{Conclusion}

This paper introduces a method named \sys for measuring knowledge similarity between DNN models. The idea is based on the insight that models with similar knowledge would group a set of inputs in similar patterns, and the decision boundaries of a model depicted by normal and adversarial input pairs are transferable to its student models. Experiments have shown that our method can achieve a high correctness on our benchmark built with popular model reuse techniques.
The source code is available at \url{https://github.com/ylimit/ModelDiff}.

%-------------------------------------------------------------------------------
\section*{Acknowledgments}
%-------------------------------------------------------------------------------

We thank all anonymous reviewers for the valuable comments. This work is done while Yunxin Liu was working at Microsoft and Ziqi Zhang was an intern at Microsoft. Ziqi Zhang contributed equally as a co-primary author. Yuanchun Li is the corresponding author.

% %-------------------------------------------------------------------------------
% \section*{Availability}
% %-------------------------------------------------------------------------------

% % USENIX program committees give extra points to submissions that are
% % backed by artifacts that are publicly available. If you made your code
% % or data available, it's worth mentioning this fact in a dedicated
% % section.

% Both the model similarity comparison tool \sys and the benchmark \bench will be open-sourced.